\documentclass[letterpaper,10pt,conference]{ieeeconf}
\IEEEoverridecommandlockouts
\usepackage{graphicx}
\usepackage{mathptmx}
\usepackage[protrusion=false]{microtype}
\usepackage{amsmath,amssymb}
\usepackage{booktabs,tabularx,multirow}
\usepackage{cite,url}
\usepackage{stfloats,placeins}
\newcommand{\sg}{\operatorname{sg}}
\newcommand{\MSE}{\operatorname{MSE}}
\title{\LARGE\bf STEP: Spatiotemporal Training with Explicit Prediction for Perceptive Humanoid Locomotion}
\author{
  Youhao Lyu$^{1*}$, Chenghao Sun$^{2*}$, Zhenxuan Ding$^{3}$, Shuting Zhang$^{4}$, Ziyang Dong$^{2}$, and Wenhao Cui$^{2\dagger}$%
  \thanks{*Equal contribution.}%
  \thanks{$^{\dagger}$Corresponding author: Wenhao Cui.}%
  \thanks{$^{1}$Youhao Lyu is with the University of Science and Technology of China (USTC), Hefei, China.}%
  \thanks{$^{2}$Chenghao Sun, Ziyang Dong, and Wenhao Cui are with Noetix Robotics, Beijing, China.}%
  \thanks{$^{3}$Zhenxuan Ding is with the Beijing University of Chemical Technology (BUCT), Beijing, China.}%
  \thanks{$^{4}$Shuting Zhang is with the National University of Defense Technology (NUDT), Changsha, China.}%
}
\begin{document}
\bstctlcite{icra2027:bstcontrol}
\maketitle
\thispagestyle{empty}
\pagestyle{empty}
\begin{abstract}

For perceptive humanoid locomotion, the relevance of terrain geometry depends on the robot's body state and how both evolve under its actions. Existing work has advanced terrain-aware sensory fusion and cross-modal prediction, improving how visual and proprioceptive information is combined and related. However, feature aggregation alone does not specify which cross-modal relationships should be learned, while prediction between simultaneous observations does not explicitly constrain their action-dependent evolution. We present Spatiotemporal Training with Explicit Prediction (STEP), a one-stage framework that jointly supervises cross-modal relationships and action-dependent state evolution within a shared control representation. Multiple learned queries aggregate depth and proprioceptive tokens into a compact policy state. Bidirectional masked prediction uses each visible modality to predict latent features of the omitted modality, explicitly supervising terrain-body relationships. An action-conditioned predictor maps the current policy state and executed action to the next fused state, with an exponential-moving-average target defined directly in the policy representation space. Simulation comparisons show overall improvements in terrain traversal over evaluated fusion alternatives and ablations, while depth-dropout and delay experiments demonstrate the benefits of both predictive objectives under degraded visual input. Deployed on a 23-DoF Noetix E1 humanoid with onboard sensing, STEP succeeds in 98 of 100 hardware trials across stairs, a 40\,cm platform, a 70\,cm gap, stepping stones, and backward stair traversal without rearward terrain visibility.
\end{abstract}

\begin{figure}[t]
\centering
\includegraphics[width=\columnwidth]{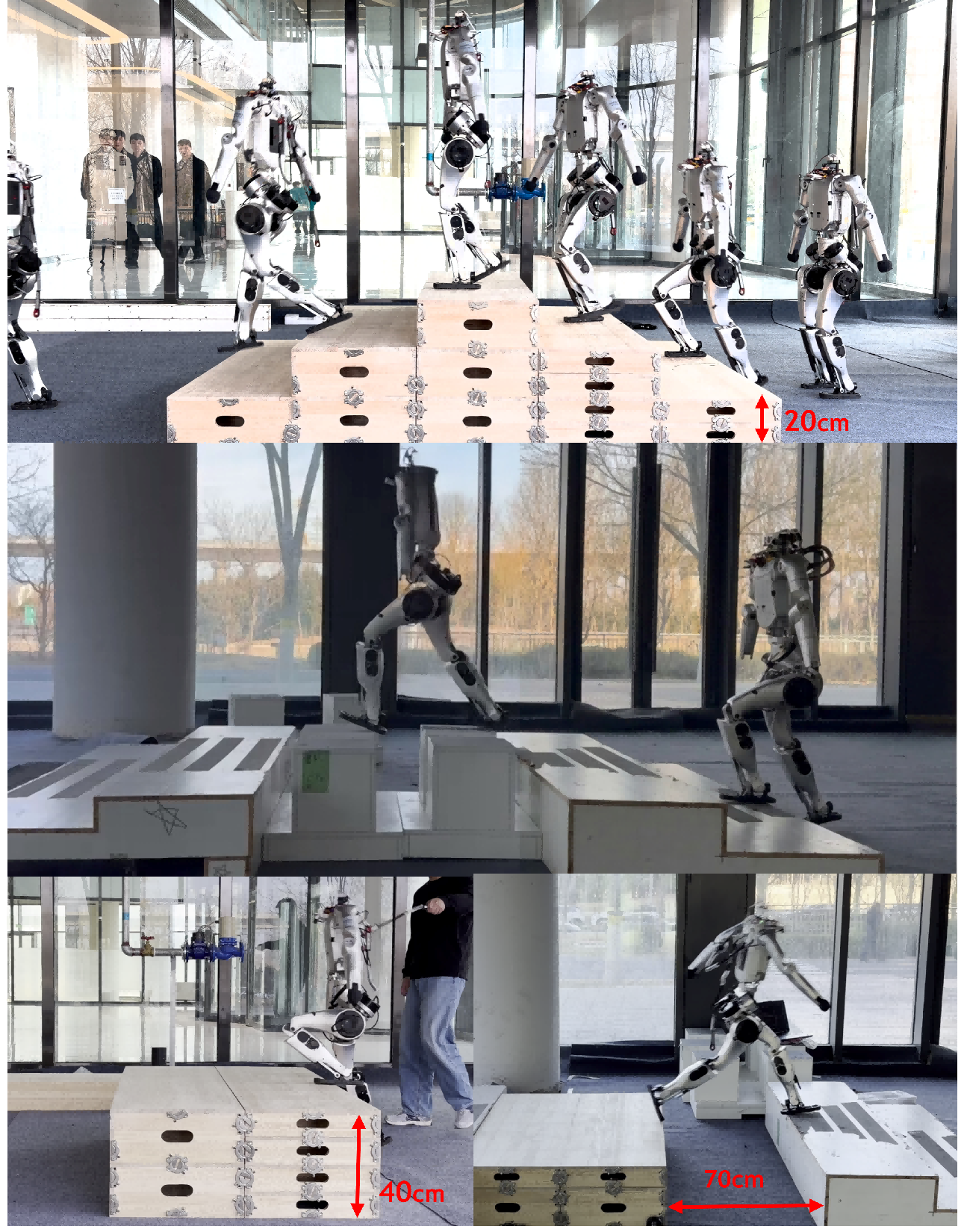}
\caption{Our STEP framework enables robust real-world humanoid locomotion, including a 40\,cm high platform, a 70\,cm gap, and 20\,cm consecutive stairs}
\label{fig:cover}
\end{figure}
\section{Introduction}
Traversing stairs, gaps, and discrete footholds requires a humanoid robot to coordinate its motion with the geometry of upcoming contacts~\cite{zhuang_humanoidparkour_2024,wang_beamdojolearning_2025}. Depth observations preview the terrain, while proprioceptive history describes body motion. The relevance of a stair edge or landing surface depends on the robot's approach velocity, configuration, and ongoing step~\cite{long_learninghumanoid_2025,miki_learningrobust_2022,cheng_extremeparkour_2024}. For the tasks in Fig.~\ref{fig:cover}, the controller must extract body--terrain information relevant to action selection and subsequent contacts while compressing heterogeneous observations into a compact state for repeated policy inference.

Proprioception-conditioned terrain aggregation selects terrain features using body-state information~\cite{he_attentionbasedmap_2025}, while cross-modal supervision provides learning signals between visual and proprioceptive observations~\cite{loquercio_learningvisual_2023}. Predictive representation learning further connects latent-state evolution to control through auxiliary prediction or model-based planning~\cite{schwarzer_spr_2021,hansen_tdmpc_2022}. These approaches provide complementary starting points, but information access and predictive supervision serve different roles: aggregation determines how observations are read, whereas prediction constrains relationships across modalities or time. For a compact locomotion state, the design challenge is to coordinate efficient information extraction with supervision of both current body--terrain relationships and action-dependent change, so that the resulting representation supports closed-loop traversal.

How can a compact control state retain relevant terrain and body information, represent their relationships and subsequent evolution, and remain efficient to compute? Our working hypothesis is that joint aggregation through multiple learned queries can provide useful information-reading paths, while cross-modal and action-conditioned temporal prediction can jointly supervise the relationships represented in the policy state. We examine this hypothesis through performance--cost comparisons and fixed-architecture objective ablations, using terrain traversal to assess the control value of the learned representation.

We present Spatiotemporal Training with Explicit Prediction (STEP), a one-stage framework that jointly learns a multimodal control state and a humanoid locomotion policy. STEP uses multiple queries to read a shared proprioceptive--depth context and unifies prediction across modalities and time: bidirectional masked prediction supervises same-time body--terrain relationships, while action-conditioned prediction supervises their subsequent fused representation. Both objectives train the policy's shared encoder jointly with proximal policy optimization (PPO)~\cite{schulman_proximalpolicy_2017}, while Sketched Isotropic Gaussian Regularization (SIGReg)~\cite{balestriero_lejepa_2025} is used to prevent latent collapse. The temporal target is the next unprojected exponential-moving-average (EMA) fused state, and prediction heads and target encoders are omitted at deployment. Simulation system comparisons show overall traversal gains, while the aggregation comparison shows higher success with slightly lower inference latency than proprioception-queried cross-attention. Hardware trials demonstrate multi-terrain locomotion, including backward stair traversal without rearward terrain visibility.

\begingroup
\setlength{\parskip}{0pt}
Our main contributions are fourfold:
\begin{itemize}[\setlength{\topsep}{2pt}]
\item \textbf{One-Stage Perceptive Locomotion Framework.} We propose STEP, a framework that jointly learns a depth-based humanoid locomotion policy and its multimodal control representation in a single policy-training process, without separate representation pre-training. Predictive supervision is confined to training, preserving direct policy inference at deployment.
\item \textbf{Multi-Query Multimodal Aggregation.} We use multiple learned queries to jointly aggregate proprioceptive and depth tokens from a shared body--terrain context. Pooling the query outputs yields a compact policy state, enabling multiple context-dependent aggregation paths within a lightweight fusion module.
\item \textbf{Spatiotemporal Multimodal Prediction.} We formulate cross-modal and action-conditioned temporal prediction as complementary supervision of a shared policy encoder: bidirectional masked prediction supervises same-time body--terrain relationships, while one-step prediction targets their subsequent fused state in the unprojected EMA representation space. Together, these objectives connect relationships across modalities and time to the representation used for control.
\item \textbf{Simulation and Real-World Validation.} We evaluate STEP through simulation system comparisons, aggregation analyses, and fixed-architecture objective ablations under nominal and degraded vision. Repeated onboard hardware trials demonstrate sim-to-real transfer across stairs, platforms, gaps, and stepping stones, including backward stair traversal without rearward terrain visibility.
\end{itemize}
\endgroup

\section{Related Work}
\subsection{Perceptive Locomotion}
Perceptive locomotion uses terrain observations to anticipate contacts and coordinate body motion over uneven ground. Maps and visual features support uneven-terrain control~\cite{hoeller_neuralscene_2022,miki_learningrobust_2022,cheng_extremeparkour_2024}, including humanoid obstacle traversal and sparse footholds~\cite{zhuang_humanoidparkour_2024,wang_beamdojolearning_2025}. PIM incorporates perception into internal-state learning~\cite{long_learninghumanoid_2025}. For terrain selection, AME/AME-2 use attentive map encoding~\cite{he_attentionbasedmap_2025,zhang_attentionneuralmap_2026}, while TAGA learns active gaze~\cite{li_taga_2026}. CReF combines proprioception-queried depth attention with recurrent integration~\cite{hao_cref_2026}. Fusion interfaces improve information access but alone do not specify supervision of cross-modal relationships or action-dependent change. STEP couples multiple learned queries over a shared proprioceptive--depth context with cross-modal and temporal prediction during policy learning.

\subsection{Multimodal Representation Learning}
Multimodal representation learning encodes complementary information from heterogeneous sensory streams for downstream robot tasks. CMS uses time-shifted proprioception to supervise visual terrain prediction~\cite{loquercio_learningvisual_2023}. RPT predicts masked image, proprioceptive, and action content during sensorimotor pre-training~\cite{radosavovic_sensorimotor_2023}. Kepler jointly encodes vision, proprioception, and force/torque using learned-query attention and masked cross-modal prediction~\cite{singh_kepler_2026}, with LeJEPA's SIGReg used to prevent latent collapse~\cite{balestriero_lejepa_2025}. RPT's separate pre-training and Kepler's single-timestep evaluation leave their application to jointly learned locomotion states unresolved. STEP adapts learned-query aggregation and bidirectional masked prediction to policy learning, jointly optimizing the encoder with PPO and temporal supervision without separate representation pre-training.

\subsection{Temporal Predictive Learning for Control}
Temporal predictive learning uses successor observations to supervise representations of robot state and its evolution. The Velocity Estimator of Long et al. combines velocity estimation with contrastive successor-state supervision of proprioceptive representations~\cite{long_hybridinternal_2024}. SPR uses a learned transition model for multi-step latent prediction against EMA targets during model-free RL~\cite{schwarzer_spr_2021}. For planning, V-JEPA 2 adds action-conditioned world-model post-training after action-free video learning~\cite{assran_vjepa2_2025}, while TD-MPC learns latent dynamics~\cite{hansen_tdmpc_2022}. These approaches leave open how temporal and cross-modal supervision can jointly shape a compact locomotion state. STEP combines cross-modal supervision with action-conditioned one-step prediction of the next unprojected EMA fused state. Prediction is training-only: deployment uses current observations and the fused state without latent rollouts or planning.

\section{Method}
\label{sec:method}

\begin{figure*}[!tb]
    \centering
    \includegraphics[width=\textwidth]{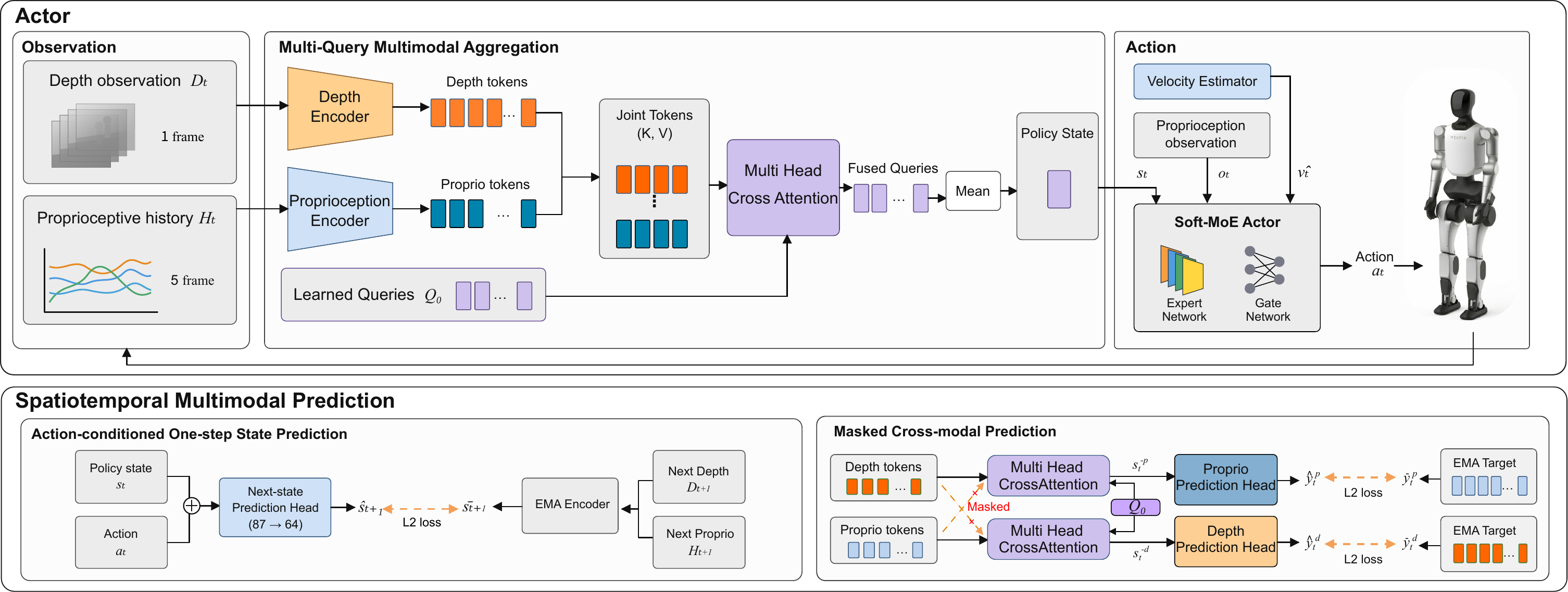}
    \caption{STEP framework. Four learned queries aggregate proprioceptive and depth tokens into a 64-dimensional policy state. The Velocity Estimator supplies detached velocity to the Soft-MoE actor. The lower row shows training-only action-conditioned prediction of the next unprojected EMA state and bidirectional masked cross-modal prediction. SIGReg (not shown) provides anti-collapse regularization. These objectives train the shared encoder jointly with PPO. Prediction heads and EMA teachers are omitted at deployment. \(\sg\) denotes stop-gradient.}
    \label{fig:system}
\end{figure*}

\subsection{Framework Overview}
\label{sec:framework}
\label{sec:implementation}

\textbf{Overall design.} Fig.~\ref{fig:system} presents an overview of STEP, an asymmetric actor--critic framework for perceptive humanoid locomotion. STEP jointly learns a multimodal control representation and a policy without separate representation pre-training. A proprioceptive branch encodes observation history, while a visual branch extracts terrain features from depth observations. Four learned queries aggregate the two modalities into a shared control state. A separately optimized Velocity Estimator provides body-frame linear velocity, and a dense Soft-MoE actor combines this estimate with the fused state and current proprioception to produce actions. The critic uses privileged information only during training.

\textbf{Policy interface.} At each control step of \(\Delta t=0.02\)\,s, the encoder \(E\) maps a five-frame proprioceptive history \(H_t\in\mathbb R^{5\times77}\) and a processed depth image \(D_t\in\mathbb R^{16\times24}\) to a fused state \(s_t\in\mathbb R^{64}\). The actor combines this state with the current proprioceptive observation \(o_t^p\in\mathbb R^{77}\) and a body-frame linear velocity estimate \(\hat v_t\in\mathbb R^3\):
\begin{equation}
    x_t=[o_t^p,\sg(\hat v_t),s_t]\in\mathbb R^{144}
    \label{eq:actor-input}
\end{equation}
Here, \(\sg\) denotes stop-gradient. A separately optimized Velocity Estimator~\cite{long_hybridinternal_2024} supplies only the detached velocity to the actor and does not participate in fusion. A dense Soft-MoE actor with four experts outputs a 23-dimensional action. Training uses an asymmetric privileged critic and task rewards combined with an AMP motion prior~\cite{peng_amp_2021}.

\textbf{Observation preprocessing.} Each 77-dimensional frame contains commands (3), body-frame angular velocity (3), roll/pitch (2), relative joint positions (23), joint velocities (23), and the previous action (23). Joint velocities are clipped to \(\pm360\) and scaled by 0.05. Five frames spaced by 0.02\,s span 0.08\,s. Depth is image-plane depth: the current noise-processing path sets values outside \([0.3,2.0]\)\,m to zero before subtracting one, making invalid values \(-1\). The camera period is 0.015\,s, distinct from the 0.02\,s control period; synchronized updates are not assumed.

\textbf{Network and training settings.} Table~\ref{tab:implementation} summarizes the network and optimization configuration.

\begin{table}[htbp]
\centering
\caption{Network and optimization settings.}
\label{tab:implementation}
\footnotesize
\setlength{\tabcolsep}{3pt}
\begin{tabularx}{\columnwidth}{@{}lX@{}}
\toprule
Component & Configuration \\
\midrule
Depth tokens & \(8\times12\) grid, width 32, linearly lifted to 64. \\
Query fuser & Four queries and two residual cross-attention blocks with four heads each, layer normalization, and feedforward sublayers. No null token. \\
Temporal predictor & \(87\!\rightarrow\!64\!\rightarrow\!64\), GELU activation. \\
Actor & Dense soft mixture of four experts, all contributing through learned gate weights. \\
PPO & 24 rollout steps per environment, five epochs, and four minibatches per epoch. Value coefficient 1, entropy coefficient 0.01, and clipping threshold 0.2 \\
SIGReg & 256 random directions, 17 integration points over \([0,3]\), and direction chunks of 64. \\
\bottomrule
\end{tabularx}
\end{table}

We sample up to 64 observations per PPO minibatch to compute the masked losses and regularization. Temporal updates independently sample up to 64 valid pairs from at most 256 environments within the current rollout.

\textbf{Velocity Estimator.} From the five-frame history \(H_t\), the estimator predicts body-frame velocity \(\hat v_t\) in m/s and a unit-normalized latent \(z_t\in\mathbb R^{16}\). A predictor uses \([z_t,a_t]\) to predict the next latent, supervised by the estimator's independent EMA encoder:
\begin{equation}
\begin{split}
\mathcal L_{\mathrm{VE}}={}&\MSE(\hat v_t,v_t^{\mathrm{critic}})\\
&+\mathbb E_t\!\left[\left\|\hat z_{t+1}-\sg(\bar z_{t+1})\right\|_2^2\right].
\end{split}
\label{eq:estimator-loss}
\end{equation}
Here, \(\MSE\) denotes mean squared error over target coordinates. Velocity supervision uses the current critic's privileged velocity, with MSE averaged over samples and three coordinates. Predicted and target latents are L2-normalized.

\textbf{Reward specification.} Task rewards and the AMP motion prior~\cite{peng_amp_2021} combine as
\begin{equation}
\begin{aligned}
    r_t^{\mathrm{task}}&=0.02\sum_k w_k f_k(t),\\
    r_t&=0.8r_t^{\mathrm{task}}+0.16\left[e^{\tanh(0.3d_t)}-e^{-1}\right],
\end{aligned}
\label{eq:reward-composition}
\end{equation}
where \(d_t\) is the AMP discriminator logit. The style term is neither scaled by the control period nor reward-normalized, whereas AMP inputs remain normalized. Table~\ref{tab:reward_terms} lists task terms and weights.

\begin{table}[htbp]
\centering
\setlength{\tabcolsep}{2pt}
\caption{Reward Term}
\label{tab:reward_terms}
\renewcommand{\arraystretch}{1.15}
\fontsize{7.5}{9}\selectfont
\begin{tabularx}{\columnwidth}{@{}>{\raggedright\arraybackslash}p{0.28\columnwidth}>{\raggedright\arraybackslash}Xr@{}}
\toprule
\textbf{Reward Term} & \textbf{Formula} & \textbf{Weight} \\
\midrule
Linear velocity & $\exp(-\lVert\mathbf u_{xy}-\mathbf v_{xy}^{\mathrm{yaw}}\rVert_2^2/0.5^2)$ & $1.2$ \\ % source: track_lin_vel_xy_exp
Yaw rate tracking & $\exp(-(u_{\mathrm{yaw}}-\omega_z^w)^2/0.5^2)$ & $1.0$ \\ % source: track_ang_vel_z_exp
Yaw rate error & $e_\omega\mathbb I[e_\omega>0.1]$\newline $\times\mathbb I[e_\omega>0.5|u_{\mathrm{yaw}}|]$ & $-0.5$ \\ % source: delta_yaw
Velocity direction & $(1-\langle\mathbf d_v,\mathbf d_u\rangle)$\newline $\times\mathbb I[\lVert\mathbf u_{xy}\rVert_2>0.1]$ & $-0.5$ \\ % source: lin_vel_dir_reward
Base orientation & $\exp(-\lVert\mathbf g_{xy}^{b}\rVert_2/0.5^2)$ & $0.5$ \\ % source: flat_orientation_exp
\midrule
Foot contact & $\mathbb I[S_{0.5}\lor(T_{\mathrm{both}}<0.2)$\newline $\qquad\lor(C<0.1)]$ & $0.5$ \\ % source: feet_contact
Foot sliding & $\sum_{f\in\{L,R\}}I_f[n(t);1]\lVert\mathbf v^w_{f,xy}\rVert_2$ & $-0.2$ \\ % source: feet_slide
Foot impact & $\mathbb I[\exists f\in\{L,R\}:$\newline $\quad\lVert\mathbf F^w_{f,xy}\rVert_2>0.5|F^w_{f,z}|]$ & $-3.0$ \\ % source: feet_impact
Foot air time & $\min(\min_{f\in\{L,R\}}m_f,0.3)$\newline $\times\mathbb I[S_{10}]\mathbb I[C>0.1]$ & $1.0$ \\ % source: feet_air_time
Foot separation & $[0.15-\delta_y]_+$ & $-1.0$ \\ % source: feet_too_near
Foot orientation & $M_A\sum_{f\in\{L,R\}}I_f[n(t);1]\lVert\mathbf g_{xy}^{f}\rVert_2$ & $-0.4$ \\ % source: feet_flat_ori
Body contact & $\sum_{b\in\mathcal B_{\mathrm{nonfoot}}} I_b[n(t);1]$ & $-10$ \\ % source: undesired_contacts
\midrule
Volume penetration & $M_A\sum_{p\in\mathcal P_t}\mathbb I[d_p>0]$\newline $\qquad\times(\lVert\mathbf v_p^w\rVert_2+10^{-6})d_p$ & $-1.0$ \\ % source: volume_points_penetration
\midrule
Leg posture & $\mathbb I[C_\perp\leq0.1]\mathbb I[C\geq0.1]$\newline $\times\sum_{j\in\mathcal J_{\mathrm{leg}}}|q_j-q_j^0|$ & $-1.0$ \\ % source: joint_deviation_leg
Upper body posture & $\sum_{j\in\mathcal J_{\mathrm{body}}}|q_j-q_j^0|$ & $-0.3$ \\ % source: joint_deviation_upbody
Standing posture & $\mathbb I[C<0.1]\sum_{j\in\mathcal J_{\mathrm{stand}}}|q_j-q_j^0|$ & $-1.0$ \\ % source: joint_deviation_standing
Joint limits & $\sum_{j\in\mathcal J_{\mathrm{all}}}([q_j^{\min,\mathrm{soft}}-q_j]_+$\newline $\qquad+[q_j-q_j^{\max,\mathrm{soft}}]_+)$ & $-10.0$ \\ % source: joint_pos_limits
Joint power & $\lVert |\boldsymbol{\tau}\odot\dot{\mathbf{q}}|\rVert_2$ & $-10^{-4}$ \\ % source: energy
Leg torques & $\sum_{j\in\mathcal J_{\mathrm{legs12}}}\tau_j^2$ & $-1.5\times10^{-7}$ \\ % source: dof_torques_l2
Joint velocity & $\sum_{j\in\mathcal J_{\mathrm{all}}}\dot q_j^2$ & $-10^{-4}$ \\ % source: dof_vel_l2
Joint acceleration & $\sum_{j\in\mathcal J_{\mathrm{all}}}\ddot q_j^2$ & $-2.5\times10^{-7}$ \\ % source: dof_acc_l2
Action rate & $\sum_{j\in\mathcal J_{\mathrm{all}}}(a_t^j-a_{t-1}^j)^2$ & $-0.1$ \\ % source: action_rate_l2
Torque limits & $\sum_{j\in\mathcal J_{\mathrm{all}}}[|\tau_j|-0.8\tau_j^{\mathrm{lim}}]_+^2$ & $-0.01$ \\ % source: torque_limits
\bottomrule
\end{tabularx}
\end{table}

\textbf{Deployment interface.} The deployed policy comprises the Velocity Estimator, multimodal encoder, and dense Soft-MoE actor. It receives a flattened five-frame proprioceptive history of shape \([B,385]\) and a flattened processed depth image of shape \([B,384]\), and outputs raw actions \(a_t^{\mathrm{raw}}\) of shape \([B,23]\), where \(B\) is the batch size. The raw actions are clipped as \(a_t=\operatorname{clip}(a_t^{\mathrm{raw}},-18,18)\) and converted into joint-position targets:
\begin{equation}
    q_{t,j}^{\mathrm{des}}=q_j^{\mathrm{default}}+s_j^{\mathrm{act}}a_{t,j},
    \qquad s_j^{\mathrm{act}}=0.25\frac{e_j^{\mathrm{lim}}}{K_{p,j}},
    \label{eq:action-mapping}
\end{equation}
where \(q_j^{\mathrm{default}}\) is the default joint position, \(e_j^{\mathrm{lim}}\) is the joint effort limit, and \(K_{p,j}\) is the position-control gain. Additionally, deployment requires maintaining the same joint ordering, observation preprocessing, and action scaling as used during training.

\subsection{Multi-Query Multimodal Aggregation}
\label{sec:representation}

To capture body--terrain relationships relevant to locomotion, we use multiple learned queries to extract complementary information from a shared multimodal context. STEP retains body and terrain tokens until joint attention, then compresses their information into a control state.

Each history frame is embedded into 64 dimensions with learned temporal-position and modality embeddings. A depth CNN produces an \(8\times12\) token grid, whose features are linearly mapped to the same width. The five proprioceptive and 96 depth tokens form a shared key--value context. Four trainable queries \(Q_t^{(0)}=Q_0\in\mathbb R^{4\times64}\) are shared across observations and initialized independently of the inputs. Two residual cross-attention blocks update these queries from the joint context. Their pooled output defines the policy state:
\begin{equation}
    s_t=E(H_t,D_t)=\frac{1}{4}\sum_{i=1}^{4}Q_{t,i}^{(2)}
    \in\mathbb R^{64},
    \label{eq:queries}
\end{equation}
where \(Q_{t,i}^{(2)}\) is the output of query \(i\) after the two blocks. The encoder \(E\) comprises the history tokenizer, depth encoder, and query fuser.

\subsection{Spatiotemporal Multimodal Prediction}
\label{sec:prediction}
\label{sec:cross-modal}
\label{sec:next-state}
\label{sec:training}

Spatiotemporal multimodal prediction consists of cross-modal prediction (CM) and action-conditioned next-state prediction (Next). CM supervises same-time modality relationships, while Next supervises the next fused state. 

\textit{Cross-modal prediction (CM).} To learn relationships between body motion and terrain observations, we train the encoder to support prediction of each modality from the other. The full input path produces \(s_t\). Mask-depth reads only the five proprioceptive tokens to produce \(s_t^{-d}\), while mask-proprio reads only the 96 depth tokens to produce \(s_t^{-p}\). All paths share fusion weights and the input-independent queries \(Q_0\), preventing the omitted modality from entering through query initialization.

The EMA teacher uses the same-time rollout history and depth. The EMA teacher uses unmasked rollout observations with the existing observation noise and depth degradations. Its depth token grid is adaptively pooled to \(2\times2\) and projected tokenwise into \(\bar y_t^d\in\mathbb R^{4\times32}\). Its five embedded history tokens are averaged and projected into \(\bar y_t^p\in\mathbb R^{32}\). Prediction heads \(h_d\) and \(h_p\) map the masked states to the respective targets:
\begin{equation}
\begin{split}
\mathcal L_{\mathrm{CM}}=\tfrac12\mathbb E_t\bigl[
 &\MSE(h_d(s_t^{-d}),\sg(\bar y_t^d))\\
 +{}&\MSE(h_p(s_t^{-p}),\sg(\bar y_t^p))\bigr].
\end{split}
\label{eq:cm}
\end{equation}
Each MSE is normalized by its target size, and the two directions receive equal weight.

\textit{Action-conditioned next-state prediction (Next).} A predictor \(g_\phi\) takes \(s_t\) and the detached command \(a_t=\operatorname{clip}(a_t^{\mathrm{raw}},-18,18)\in\mathbb R^{23}\), passed to the environment before conversion to joint-position targets:
\begin{align}
    \hat s_{t+1}&=g_\phi([s_t,\sg(a_t)]),\\
    \bar s_{t+1}&=\bar E(H_{t+1},D_{t+1}),\\
    \mathcal L_{\mathrm{Next}}
    &=\mathbb E_{t\in\mathcal V}\!\left[
      \MSE\bigl(\hat s_{t+1},\sg(\bar s_{t+1})\bigr)\right].
    \label{eq:next-state}
\end{align}
The teacher encodes cached next-step policy history and depth to produce an unprojected 64-dimensional target in the policy's representation space.

\textit{Anti-collapse regularization.} Representation collapse is a common challenge in representation learning. Sketched Isotropic Gaussian Regularization (SIGReg)~\cite{balestriero_lejepa_2025} discourages degenerate representations by matching empirical characteristic functions along random unit directions to those of an isotropic Gaussian. It acts on the unmasked online modality projections \(y^d,y^p\), corresponding to the teacher targets, and directly on the fused state:
\begin{equation}
    \mathcal R_{\mathrm{modal}}
      =\tfrac12[\mathcal R(y^d)+\mathcal R(y^p)],
    \qquad \mathcal R_{\mathrm{joint}}=\mathcal R(s).
    \label{eq:regularization}
\end{equation}
For \(\mathcal R(y^d)\), the four 32-dimensional depth projections from each observation are pooled across the batch as separate samples.

\textit{Joint training.} The main optimizer combines PPO~\cite{schulman_proximalpolicy_2017}, including policy, value, and entropy terms, with the predictive and regularization objectives:
\begin{equation}
\begin{split}
\mathcal L_{\mathrm{main}}=\mathcal L_{\mathrm{PPO}}+\alpha_k\bigl(
 &\mathcal L_{\mathrm{Next}}+\mathcal L_{\mathrm{CM}}\\
 &+0.02\mathcal R_{\mathrm{modal}}
 +0.005\mathcal R_{\mathrm{joint}}\bigr),
\end{split}
\label{eq:total-loss}
\end{equation}
Here, \(\alpha_k=0.1\min(k/200,1)\) at iteration \(k\). After rollout collection, returns and GAE are computed with timeout value bootstrapping. Each minibatch updates the AMP discriminator, the main network, then the independently optimized Velocity Estimator and its EMA target, in that order. Rollout AMP rewards are not recomputed. Neither fusion nor estimator losses enter the scalar reward. The fusion teacher is initialized from the online network, frozen in evaluation mode throughout each PPO iteration, and then updated once:
\begin{equation}
    \bar\theta\leftarrow0.99\bar\theta+0.01\theta,
    \label{eq:ema-update}
\end{equation}
with registered buffers copied directly. The estimator's independent target uses the same coefficient after each estimator update. The registered fusion-teacher EMA set covers the CNN, fuser, modality heads, normalizer, and Estimator, exceeding the fused-state computation. Checkpoints restore teacher weights without reinitialization.

\begin{table*}[!htb]
\centering
\caption{Baseline comparisons and ablations on multi-terrain success rate (SR, \%) and traversal rate (TR, \%). Bold indicates the best value in each column. Overall Mean averages all 16 reported SR/TR entries.}
\label{tab:main}
\resizebox{\textwidth}{!}{%
\begin{tabular}{l*{17}{c}}
\toprule
\multirow{3}{*}{Method} & \multicolumn{4}{c}{Gap} & \multicolumn{4}{c}{Stairs Up} & \multicolumn{4}{c}{Stairs Down} & \multicolumn{4}{c}{Stepping Stones} & \multirow{3}{*}{\shortstack{Overall\\Mean}} \\
\cmidrule(lr){2-5} \cmidrule(lr){6-9} \cmidrule(lr){10-13} \cmidrule(lr){14-17}
& \multicolumn{2}{c}{L5} & \multicolumn{2}{c}{L8} & \multicolumn{2}{c}{L5} & \multicolumn{2}{c}{L8} & \multicolumn{2}{c}{L5} & \multicolumn{2}{c}{L8} & \multicolumn{2}{c}{L5} & \multicolumn{2}{c}{L8} & \\
\cmidrule(lr){2-3} \cmidrule(lr){4-5} \cmidrule(lr){6-7} \cmidrule(lr){8-9} \cmidrule(lr){10-11} \cmidrule(lr){12-13} \cmidrule(lr){14-15} \cmidrule(lr){16-17}
& SR & TR & SR & TR & SR & TR & SR & TR & SR & TR & SR & TR & SR & TR & SR & TR & \\
\midrule
STEP & \textbf{100.00} & \textbf{100.00} & \textbf{99.41} & \textbf{99.71}
& \textbf{99.76} & \textbf{99.89} & \textbf{90.92} & 92.95
& \textbf{99.76} & \textbf{99.98} & \textbf{95.46} & \textbf{98.73}
& \textbf{93.80} & \textbf{97.25} & \textbf{74.90} & \textbf{80.18}
& \textbf{95.17} \\

w/o Prediction & \textbf{100.00} & \textbf{100.00} & 99.22 & 99.59
& 99.71 & 99.77 & 88.33 & \textbf{93.07}
& 99.56 & 99.84 & 81.93 & 93.05
& 85.06 & 88.80 & 57.23 & 71.18
& 91.02 \\

w/o Multi Query & 91.41 & 91.63 & 83.69 & 87.83
& 89.40 & 91.72 & 74.27 & 85.86
& 88.38 & 92.72 & 62.89 & 68.39
& 33.20 & 37.48 & 17.87 & 22.95
& 69.98 \\

Proprio-Query & 99.66 & 99.80 & 99.02 & 99.37
& 99.37 & 99.67 & 79.25 & 83.38
& 99.41 & 99.63 & 80.13 & 85.51
& 75.15 & 79.35 & 43.51 & 56.53
& 86.17 \\
\bottomrule
\end{tabular}%
}
\end{table*}
\section{Experiments}
\label{sec:experiments}
We evaluate STEP through four sets of experiments. Section~\ref{sec:locomotion} compares terrain-traversal performance against reference and ablated configurations. Section~\ref{sec:aggregation} analyzes fusion efficiency, query contributions, and spatial attention. Section~\ref{sec:c2} evaluates predictive objectives under degraded visual input while keeping the deployed fusion architecture fixed. Finally, Sec.~\ref{sec:c3} examines real-world locomotion across multiple terrains. Unless otherwise stated, all STEP-based variants share the same task reward, style reward, and curriculum.

All policies are trained in \emph{Isaac Lab}~\cite{nvidia_isaaclab_2025} using PPO with the RSL-RL training stack~\cite{schulman_proximalpolicy_2017,schwarke_rslrllearning_2025}. Each configuration is evaluated using 2048 parallel simulation environments.

Table~\ref{tab:terrain-curriculum} lists the parameter ranges of six terrain types used in the training curriculum.

\begin{table}[!ht]
\centering
\caption{Terrain parameter ranges (m), before discretization.}
\label{tab:terrain-curriculum}
\footnotesize
\setlength{\tabcolsep}{3pt}
\begin{tabularx}{\columnwidth}{@{}l>{\raggedright\arraybackslash}X@{}}
\toprule
Terrain & Parameter ranges\\
\midrule
Rough flat & Height noise: 0.01--0.05.\\
Platform & Obstacle height: 0.20--0.40.\\
Stairs up & Step height: 0.05--0.25, width: 0.25--0.50.\\
Stairs down & Step height: 0.05--0.25, width: 0.25--0.50.\\
Gap & Width: 0.15--0.75; depth: 0.20--1.00, support-height variation scale: 0.03--0.10.\\
Stepping stones & Stone width: 0.25--0.50, spacing: 0.10--0.35, pit floor: $[-0.50,-0.30]$.\\
\bottomrule
\end{tabularx}
\end{table}

\subsection{Terrain Traversability Analysis}
\label{sec:locomotion}

\subsubsection{Experimental Setup}
We evaluate terrain traversal on gaps, stair ascent, stair descent, and stepping stones at curriculum levels L5 and L8. Table~\ref{tab:main} reports \textbf{success rate (SR)} and \textbf{traversal rate (TR)}. The target traversal distance is half the length of the corresponding terrain map. SR measures the fraction of trials that reach this target, while TR normalizes the traversed distance before termination by the same target distance. Both metrics are reported as percentages. Overall Mean averages the 16 SR/TR entries, summarizing completion and partial progress across terrain settings.

\subsubsection{Method}
We compare STEP with two internal ablations and a reference implementation (Table~\ref{tab:main}). \textbf{w/o Prediction} removes predictive objectives while retaining multi-query aggregation. \textbf{w/o Multi Query} reduces the number of learned aggregation queries from four to one while retaining the predictive objectives. \textbf{Proprio-Query} uses proprioceptive features to query depth features through cross-attention, implemented with reference to AME-2~\cite{zhang_attentionneuralmap_2026} and CReF~\cite{hao_cref_2026}.

\subsubsection{Results and Analysis}
STEP achieves the highest Overall Mean of 95.17\% (Table~\ref{tab:main}), with near-saturated performance on gaps and easier stairs and clearer gains on difficult stair descent and stepping stones. Both STEP and w/o Prediction outperform Proprio-Query overall, supporting the proposed aggregation design even without auxiliary representation-learning objectives. The full framework provides additional gains, although it does not lead on every metric.

Compared with STEP, w/o Prediction deteriorates notably on stair descent and stepping stones, especially at the higher difficulty level. This supports the combined contribution of predictive supervision on challenging terrain. Despite retaining the predictive objectives, w/o Multi Query performs worse than STEP, particularly on stepping stones. This comparison supports the benefit of using four learned aggregation queries rather than one in the evaluated configuration.

\begin{figure*}[t]
\centering
\begin{minipage}[t]{0.43\textwidth}
    \centering
    \includegraphics[width=\linewidth,height=1.85in,keepaspectratio]{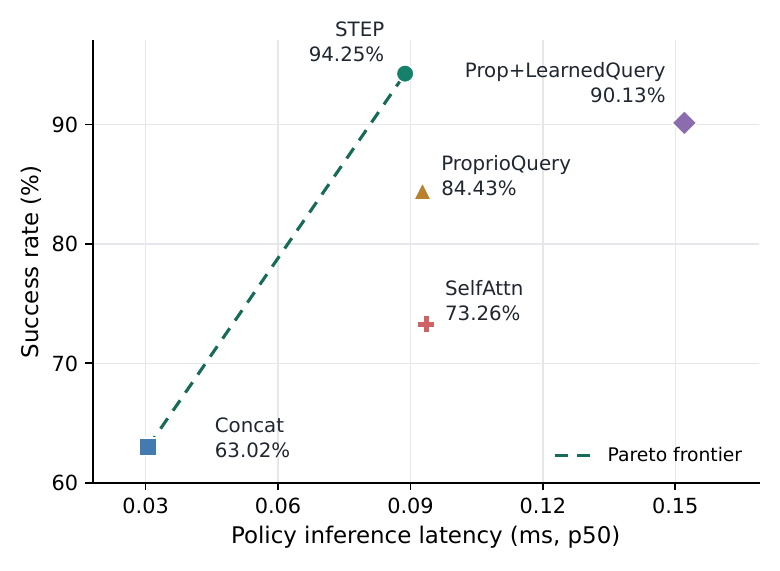}
    \par\footnotesize (a) Success--latency trade-off
\end{minipage}\hfill
\begin{minipage}[t]{0.55\textwidth}
    \centering
    \includegraphics[width=\linewidth,height=1.85in,keepaspectratio]{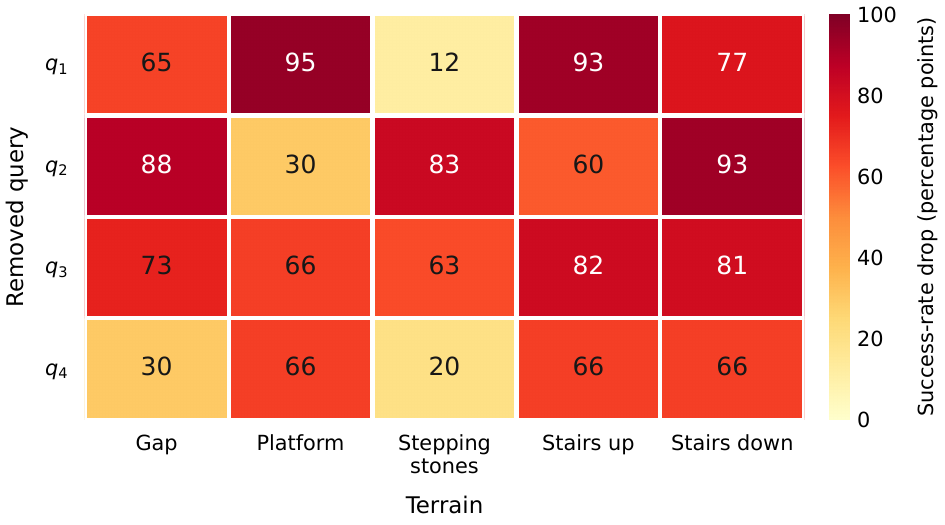}
    \par\footnotesize (b) Query-removal effects
\end{minipage}
\par\smallskip
\includegraphics[width=0.85\textwidth,trim=65bp 195bp 154bp 170bp,clip]{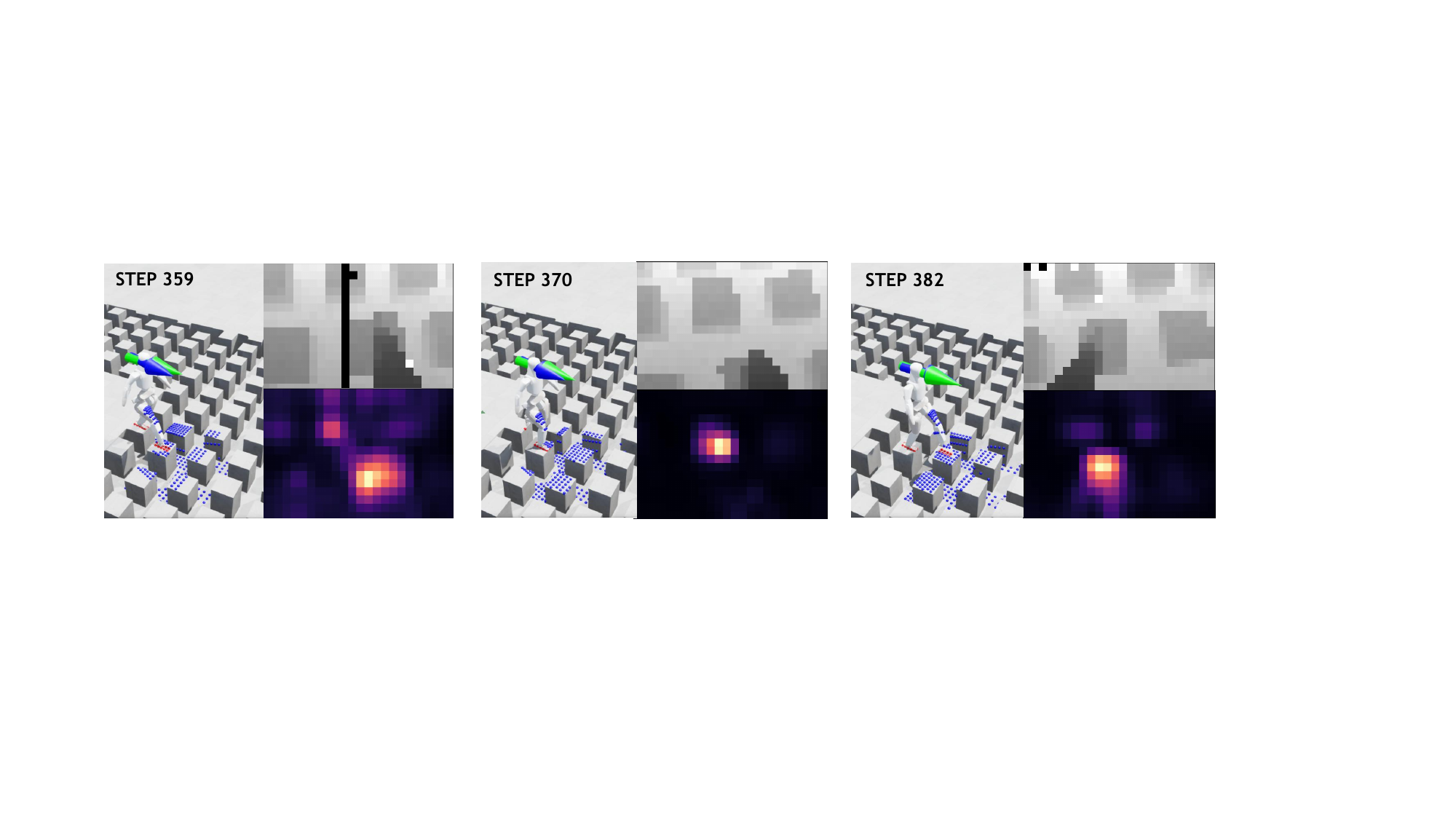}
\par\footnotesize (c) Depth attention during stepping-stone traversal
\caption{Attentive fusion analysis. (a) Success rate versus median policy-only inference latency (p50). The dashed line marks the Pareto frontier. (b) Success-rate decrease (percentage points) after removing each query across terrains. (c) Robot views, depth observations, and attention maps at steps 359, 370, and 382 show changing spatial focus.}
\label{fig:query-analysis}
\end{figure*}

\subsection{Control-Conditioned Attentive Fusion Ablation}
\label{sec:aggregation}

\subsubsection{Experimental Setup}
We examine the aggregation design through three complementary analyses: the success--latency trade-off across fusion configurations, the effect of removing individual queries, and the spatial distribution of depth attention (Fig.~\ref{fig:query-analysis}).

\textbf{Fusion efficiency.} We compare Concat, ProprioQuery, SelfAttn, LearnedQuery, and STEP under the same training objectives. We select traversal success rate (SR, \%) against median policy inference latency (p50, ms) as a benchmark, separating control performance from inference cost. The latency metric concerns policy inference rather than the end-to-end sensing and control loop. The plotted Pareto frontier identifies configurations for which no other evaluated configuration has both at least as high an SR and at most as much latency, with a strict improvement in at least one metric.

\textbf{Query contributions.} We assess the four aggregation queries separately on gaps, platforms, stair ascent, stair descent, and stepping stones. For each terrain, we record the SR of the intact configuration minus the SR with query \(q_i\) removed, for \(i\in\{1,2,3,4\}\). Differences are expressed in percentage points, not relative percentages. A larger positive value indicates greater performance degradation under that query intervention. Reporting each terrain separately exposes differences in query reliance that an overall score could obscure.

\subsubsection{Results and Analysis}
Figure~\ref{fig:query-analysis}(a) places STEP on the plotted Pareto frontier alongside Concat: STEP achieves the highest success rate, while Concat has the lowest latency. This trade-off favors STEP when traversal performance is prioritized, without making it the minimum-cost option. 

Query removal reveals terrain-dependent reliance on the aggregation paths (Fig.~\ref{fig:query-analysis}(b)): \(q_1\) contributes most strongly to platforms and stair ascent, whereas \(q_2\) is particularly important for stepping stones and stair descent. The attention maps also shift between stepping-surface regions as the robot advances (Fig.~\ref{fig:query-analysis}(c)). Together, these observations suggest differentiated, context-dependent information selection, without assigning a unique semantic role to each query.

\begin{figure*}[!t]
\centering
\includegraphics[width=0.98\textwidth,trim=50bp 95bp 110bp 40bp,clip]{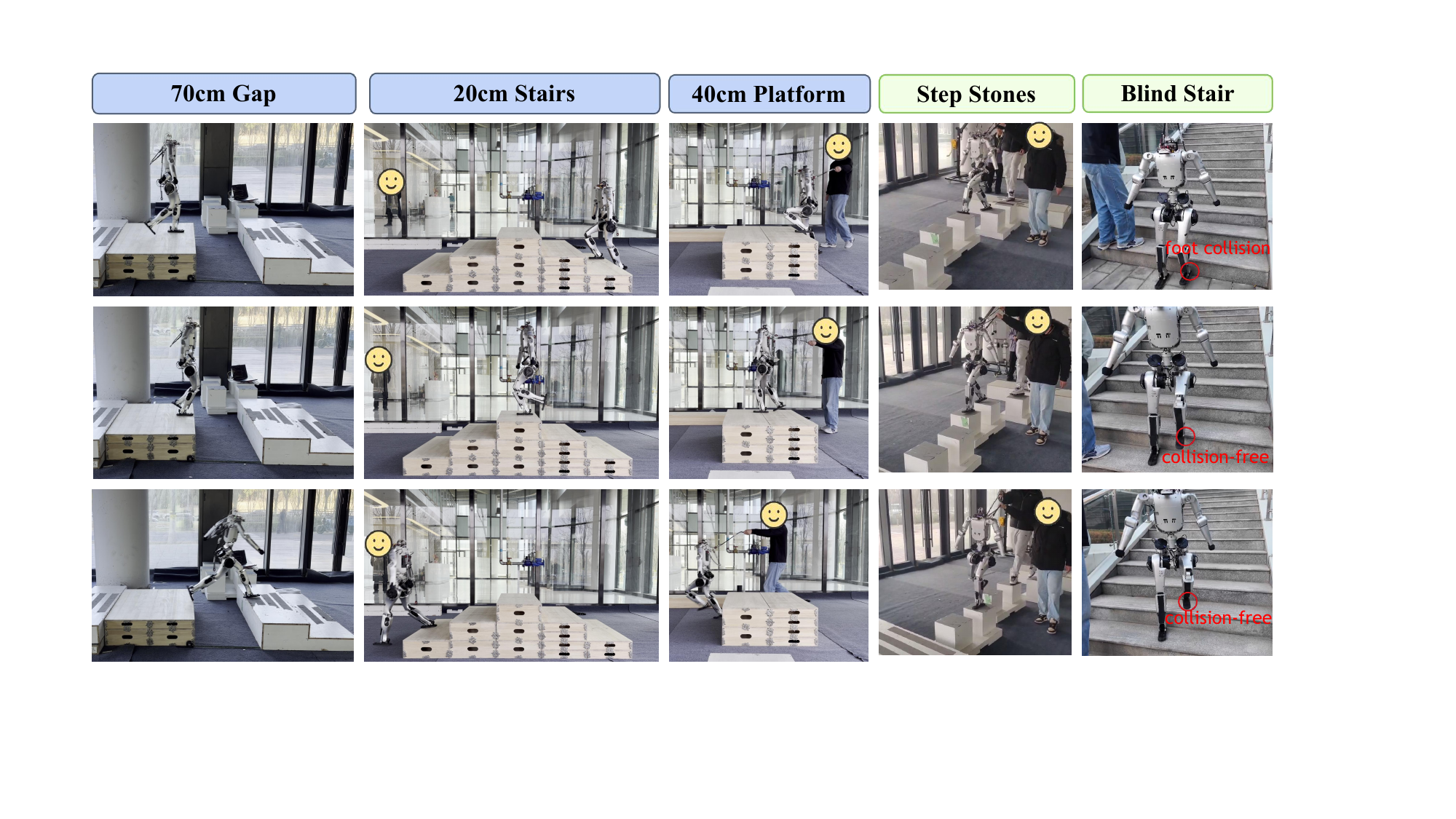}
\caption{Real-world traversal sequences of STEP on a gap, stairs, a platform, stepping stones, and stairs with limited visual coverage. Each column shows successive stages of traversal. }
\label{fig:legacy-hardware}
\end{figure*}

\subsection{Spatiotemporal Multimodal Prediction Ablation}
\label{sec:c2}

\subsubsection{Experimental Setup}
We test two separate visual perturbations on gaps, stair ascent, stair descent, stepping stones, and high platforms at curriculum level L6. \textbf{Depth dropout} disables depth updates throughout evaluation, holding the last available frame without resuming updates. \textbf{Depth delay} uses the depth input from the preceding control step, introducing a one-step delay at the 20\,ms control period. Terrain conditions and perturbation settings are shared across methods.

\subsubsection{Method}
We compare STEP with three ablations. The \textbf{w/o CM} and \textbf{w/o Next} variants remove cross-modal prediction (CM) and action-conditioned next-state prediction (Next), respectively. \textbf{Fusion-only} disables all representation-learning auxiliaries. All four variants retain the same four-query fusion architecture and control components. Table~\ref{tab:real_exp} reports SR.

\begin{table*}[t]
\centering
\caption{Visual-degradation comparison across five terrains at L6 under continuous depth dropout and a one-control-step (20\,ms) depth delay. Entries report success rate (SR, \%).}
\label{tab:real_exp}
\fontsize{6.966}{8.4}\selectfont
\setlength{\tabcolsep}{3pt}
\begin{tabular*}{\textwidth}{@{\extracolsep{\fill}}l*{10}{c}@{}}
\toprule
\multirow{2}{*}{Method} & \multicolumn{2}{c}{Gap} & \multicolumn{2}{c}{Stairs Up} & \multicolumn{2}{c}{Stairs Down} & \multicolumn{2}{c}{Stepping Stones} & \multicolumn{2}{c}{High Platform}\\
\cmidrule(lr){2-3}\cmidrule(lr){4-5}\cmidrule(lr){6-7}\cmidrule(lr){8-9}\cmidrule(lr){10-11}
 & \shortstack{Depth\\dropout} & \shortstack{Depth delay\\(20 ms)} & \shortstack{Depth\\dropout} & \shortstack{Depth delay\\(20 ms)} & \shortstack{Depth\\dropout} & \shortstack{Depth delay\\(20 ms)} & \shortstack{Depth\\dropout} & \shortstack{Depth delay\\(20 ms)} & \shortstack{Depth\\dropout} & \shortstack{Depth delay\\(20 ms)}\\
\midrule
\textbf{STEP}   & 0 & \textbf{96.8} & \textbf{100}  & \textbf{91.2} & \textbf{84.5} & \textbf{98.7} & 0 & \textbf{88.6} & \textbf{89.1} & \textbf{95.4} \\
w/o CM          & 0 & 23.5          & 94.2 & 88.0          & 81.2          & 65.4          & 0 & 0          & 75.8          & 71.9          \\
w/o Next   & 0 & 92.1          & 92.8 & 87.4          & 80.0          & 64.2          & 0 & 54.1          & 84.5          & 90.8          \\
Fusion-only     & 0 & 19.4          & 88.5 & 84.3          & 76.9          & 61.0          & 0 & 0         & 71.2          & 67.3          \\
\bottomrule
\end{tabular*}
\end{table*}

\subsubsection{Results and Analysis}
As shown in Table~\ref{tab:real_exp}, all methods fail on gaps and stepping stones under continuous depth dropout, suggesting that proprioceptive history and a frozen depth frame do not adequately support traversal of these discontinuous terrains in the evaluated setting. STEP nevertheless achieves the highest SR on stair ascent, stair descent, and high platforms, and leads across all terrains under depth delay. These results support its overall robustness advantage among the tested configurations while revealing a shared limitation when fresh depth observations remain unavailable.

 Under depth delay, w/o Next deteriorates markedly on stepping stones despite retaining both input modalities, supporting the value of temporal supervision beyond multimodal aggregation alone. The w/o CM variant suffers larger losses on gaps and stepping stones, with no successful stepping-stone trials. Both ablations also substantially reduce stair-descent performance. Overall, these results highlight the importance of both cross-modal and temporal prediction under unreliable visual input, with their joint supervision substantially improving policy robustness in the evaluated settings.

\subsection{Real-World Experiments}
\label{sec:c3}

\subsubsection{Experimental Setup and Method}
We evaluate STEP on the \emph{Noetix E1}, a 23-DoF humanoid with onboard IMU, joint encoders, and a base-mounted depth camera pitched at $60^{\circ}$ relative to the horizontal plane. Tasks include 20\,cm stairs, a 40\,cm platform, a 70\,cm gap, and randomly arranged stepping stones. A visual-degradation test additionally evaluates backward ascent and descent on 15\,cm stairs without rearward terrain visibility.

 Hardware evaluation tests STEP alone, with 20 trials per task and no human intervention during traversal. Success requires traversing the terrain and reaching a stable standing posture. Each backward-stair trial comprises a complete ascent followed by descent. Table~\ref{tab:hardware-results} reports completion counts.

\begin{table}[h]
\centering
\caption{Real-world traversal results of STEP.}
\label{tab:hardware-results}
\small
\begin{tabular*}{\columnwidth}{@{\extracolsep{\fill}}llc@{}}
\toprule
Task & Setting & Success / Trials\\
\midrule
Stairs & Indoor,  20\,cm & 20/20\\
Platform & Indoor, 40\,cm & 20/20\\
Gap & Indoor, 70\,cm & 18/20\\
Stepping Stones & Indoor, random & 20/20\\
Visual Degradation & 15\,cm, backward & 20/20\\
\bottomrule
\end{tabular*}
\end{table}

\subsubsection{Results and Analysis}
STEP completes 98 of 100 hardware trials (Table~\ref{tab:hardware-results}), supporting the transfer of the jointly trained policy and representation to multi-terrain locomotion with onboard sensing. Gap crossing accounts for both failures, with 18 successful completions, indicating that reliability still varies across the tested tasks.

Successful backward stair traversal demonstrates operation when upcoming terrain lies outside the camera view. The illustrated sequence shows an initial foot collision followed by collision-free steps (Fig.~\ref{fig:legacy-hardware}), suggesting that the controller can adjust its behavior with limited visual coverage, without establishing online learning or reconstruction of unseen terrain.

\section{Conclusion}
\label{sec:conclusion}
We presented STEP, a one-stage framework that combines multi-query aggregation with cross-modal and action-conditioned prediction to learn a shared multimodal control state and locomotion policy. STEP achieves 98/100 successful hardware trials using onboard sensing, including backward stair traversal without rearward terrain visibility. Sustained depth dropout remains challenging, particularly on gaps and stepping stones. 

Sustained depth dropout remains a limitation: all evaluated configurations fail on discontinuous terrains when depth updates remain unavailable. Future work will extend prediction beyond the current one-step horizon and investigate temporal memory for retaining previously observed terrain information, with the aim of improving locomotion under intermittent visual input.

\bibliographystyle{IEEEtran}
\bibliography{arxiv}
\end{document}